\documentclass[11pt]{article}

\usepackage[final]{acl}
\usepackage{booktabs}
\usepackage{amsmath}
\usepackage{amssymb}
\usepackage{times}
\usepackage{latexsym}
\usepackage{comment}
\usepackage[T1]{fontenc}

\usepackage[utf8]{inputenc}

\usepackage{microtype}

\usepackage{inconsolata}

\usepackage{graphicx}

\usepackage{listings}
\usepackage{xcolor}
\usepackage{cuted}
\usepackage{float}
\usepackage{algorithm}
\usepackage{algpseudocode}
\usepackage{threeparttable}

\lstdefinestyle{spawnlorastyle}{
    backgroundcolor=\color{white},
    commentstyle=\color{gray}\itshape,
    keywordstyle=\bfseries,
    numberstyle=\tiny\color{gray},
    basicstyle=\ttfamily\scriptsize,
    breakatwhitespace=false,
    breaklines=true,
    captionpos=t,
    keepspaces=true,
    numbers=left,
    numbersep=5pt,
    showspaces=false,
    showstringspaces=false,
    showtabs=false,
    tabsize=4,
    escapeinside={(*@}{@*)},
    frame=tb,
    rulecolor=\color{black}
}

\title{
Routing Is Not Enough: Diagnosing Intra-Adapter Subspace Contention in MoE+LoRA Fine-Tuning}

\author{
  \textbf{Mehreen Hossain Chowdhury\textsuperscript{1,*}} \quad
  \textbf{Nowshin Mahjabin\textsuperscript{1,*}} \quad
  \textbf{Ahmed Shafin Ruhan\textsuperscript{1,*}}
  \\
  \textbf{Md Azam Hossain\textsuperscript{1}} \quad
  \textbf{Abu Raihan Mostofa Kamal\textsuperscript{1}} \quad
  \textbf{Md Tahmid Rahman Laskar\textsuperscript{2,3}}
  \\
  \\
  \textsuperscript{1}Islamic University of Technology
  \\
  \textsuperscript{2}York University
  \qquad
  \textsuperscript{3}Dialpad Inc.
  \\
  \textsuperscript{*}Equal contribution.
}

\begin{document}
\maketitle
\begin{abstract}
Multi-domain fine-tuning often combines MoE routing with LoRA, assuming that token-level routing separates domain-specific updates. We test this assumption in MoE+LoRA using Python code paired with biomedical text and mathematical reasoning. Although these domains show near-disjoint expert routing, adding biomedical data substantially increases code perplexity, indicating that routing separation alone may not prevent negative transfer. To localize the failure, we introduce Jaccard routing overlap and adapter-gradient cosine similarity, measuring expert sharing and update compatibility, respectively. These diagnostics indicate that interference arises 
mostly from nearly orthogonal domain gradients competing within the same low-rank adapter subspace. We address this with SpawnLoRA, which dynamically adds gated sub-adapters inside MoE experts when adapter-level contention is detected, while keeping the router fixed. We evaluate SpawnLoRA on Phi-tiny-MoE-instruct and OLMoE-1B-7B across multiple mixture settings and find that it effectively reduces negative transfer over standard and rank-adaptive LoRA. This demonstrates that structural separation inside experts helps beyond routing or rank expansion alone.

\end{abstract}
\vspace{-2mm}
\section{Introduction}
\label{sec:intro}
\vspace{-2mm}

Parameter-efficient fine-tuning enables large language models to adapt to new domains without updating all parameters \cite{hanparameter}. Low-Rank Adaptation (LoRA) does this by inserting small trainable matrices into frozen models \citep{hu2022lora}, while Mixture-of-Experts (MoE) architectures route each token to a sparse subset of expert modules, increasing effective capacity without activating the full model \citep{shazeer2017outrageously,fedus2022switch}. Together, these ideas suggest a practical multi-domain adaptation strategy: attach LoRA adapters to MoE experts and rely on token-level routing to separate domain updates.
{\color{black}
This setting is particularly relevant when open-weight MoE models are adapted over time to handle multiple domains within a shared deployment. In practice, domain boundaries are not always clear, and individual inputs may contain information from more than one domain. Reliable domain labels may also be unavailable at inference time, making fixed domain-specific adaptation difficult to apply. This motivates adaptation methods that can specialize without depending on explicit domain identity.
}

{\color{black}
Prior work has used routing over lightweight or LoRA-based experts to encourage
specialization in parameter-efficient multi-task and instruction tuning
\citep{zadouri2024pushing,zhu2024mixlora,dou2024loramoe}.
However, it remains unclear whether routing-based specialization alone is
sufficient to prevent adapter-level negative transfer during MoE fine-tuning.} Even when expert routing succeeds, domain-specific updates may still interfere inside the shared low-rank adapters of experts.

This motivates our central question: when MoE routing already separates domains at the expert level, is that separation sufficient to prevent negative transfer in the LoRA adapters attached to those experts? We study this question in a controlled multi-domain setting where Python code serves as the primary evaluation domain and each added domain acts as a potential source of interference. We focus on Python code as the protected domain because its syntax and token structure differ sharply from natural-language text, making it a useful target for cross-domain interference. We define negative transfer as the increase in held-out code perplexity over the corresponding code-only run, pairing code with biomedical text to study a severe mismatch and mathematical reasoning to test a more moderate shift.

Our analysis reveals that while domains often follow nearly disjoint expert paths, adding a second domain can still degrade code perplexity. To explain why, we introduce two diagnostics: Jaccard routing overlap (expert sharing across domains) and adapter-gradient cosine similarity (alignment of their LoRA updates). The diagnostics show interference arises mainly from nearly orthogonal domain gradients competing inside the same low-rank adapter. We call this failure mode \textit{intra-adapter subspace contention}, where further increase in shared rank is found to be insufficient. 

Motivated by this diagnosis, we propose \textbf{SpawnLoRA} (Fig.~\ref{fig:method}), which adds gated LoRA sub-adapters inside an expert when contention is detected. We evaluate SpawnLoRA across two MoE backbones and multiple conflict levels to show its effectiveness in reducing negative transfer. Our key contributions are threefold: \textbf{(1)} we show that successful token-level routing can coexist with substantial adapter-level interference in MoE+LoRA fine-tuning, and introduce routing- and gradient-based diagnostics to localize this failure mode; \textbf{(2)} we show that expanding a shared adapter rank can worsen contention under conflict; \textbf{(3)} our proposed \textbf{SpawnLoRA} ensures structural separation inside experts to 
reduce negative transfer.

\section{Related Work}
\label{sec:related}


\noindent \textbf{Gradient conflict and capacity expansion.}

Multi-domain fine-tuning can fail when domains induce incompatible gradients in shared parameters. {\color{black}
Rank-adaptive LoRA methods dynamically adjust adaptation capacity during training
\citep{zhang2023adalora,liu2024drlora}. AdaLoRA reallocates rank based on parameter importance, while DR-LoRA periodically expands expert ranks according to expert saliency. In DR-LoRA, this expansion increases capacity within the existing adapter rather than creating separate adaptation pathways, so updates from different domains may still interact within the same shared low-rank subspace.
}

Gradient surgery methods instead modify conflicting updates
\citep{yu2020gradient,wang2021gradient}. {\color{black}
MoE-style PEFT methods use routing to encourage specialization across lightweight or LoRA-based experts
\citep{zadouri2024pushing,zhu2024mixlora}.
}
Our diagnosis shows that near-orthogonal gradients can still compete within a single low-rank adapter despite strong routing separation, suggesting that shared capacity expansion alone may be insufficient.

\noindent \textbf{Structural separation.}
{\color{black}
Other works promote structural separation through per-expert, hierarchical, or dynamically expanded adapters
\citep{zadouri2024pushing,zhu2024mixlora,peng2026hilora,huynh2025mixlora}.
}
{\color{black}
Recent MoE+LoRA methods such as MixLoRA and LoRAMoE combine LoRA-style adaptation with expert or mixture structures \citep{zhu2024mixlora,dou2024loramoe}. More generally, domain-specific adapters provide explicit separation when domain identity is known, but rely on predefined specialization rather than detecting where interference arises during training. SpawnLoRA operates inside individual MoE experts and selectively introduces ReLU-gated sub-adapters only where gradient-space contention is detected.
} 
\section{Experimental Setup}
\label{sec:setup}

\paragraph{Models.}
We use two MoE models to examine whether intra-adapter contention and SpawnLoRA's mitigation are specific to one architecture or hold across many. 
We use Phi-tiny-MoE-instruct \citep{lislimmoe} (1.1B active/3.8B total, 16 experts, top-2 routing) and OLMoE-1B-7B \citep{muennighoff2024olmoe} (1B active/7B total, 64 experts, top-8 routing).

\paragraph{Data.}
We use Python code from HumanEval \citep{chen2021evaluatinglargelanguagemodels} as the anchor domain and measure negative transfer through held-out code perplexity. {\color{black}
We compute code perplexity using standard next-token prediction, with cross-entropy loss accumulated over all predicted tokens. We then pool the total loss and token count across the evaluation corpus and compute $\mathrm{PPL}=\exp(\mathrm{total\ loss}/\mathrm{total\ tokens})$. 
}

We pair code with two interfering domains: PubMedQA biomedical text \citep{jin2019pubmedqadatasetbiomedicalresearch}, representing a severe domain mismatch, and GSM8K mathematical reasoning \citep{cobbe2021trainingverifierssolvemath}, representing a more moderate shift. This two-pair design tests whether intra-adapter contention arises only under strong domain mismatch or also under closer domain pairs. For each domain pair, we use a fixed training budget of 2,000 examples and 3,000 steps. We conduct experiments across 3 runs with different data mixture ratios. Run A uses code only and serves as the single-domain reference; Run B uses an 80/20 code/interfering-domain mixture; and Run C uses a 50/50 mixture. {\color{black} 
To account for differences in code exposure across the mixture settings, we also use code-only controls matched to Runs B and C in both unique-code coverage and number of code updates. The Run B control uses 1,600 code examples over 2,400 updates, while the Run C control uses 1,000 examples over 1,500 updates. We use these controls to estimate how much of the observed degradation could be explained by reduced code exposure alone.
}

\paragraph{Scope.}
Our setup is deliberately small-scale (1--8B active parameters, 3,000 steps) to induce and isolate a measurable contention signal rather than maximize task performance. Absolute code perplexity is therefore not comparable to converged models; we report \textit{negative transfer} as a within-method change relative to Run~A, which 
isolates the effect of domain interference. 

\section{Diagnostic Study}
\label{sec:diag}
We first diagnose why multi-domain MoE+LoRA fine-tuning can fail and explore the following research question: \textit{If there are any negative transfers, do they come from routing overlap or from conflict inside the LoRA adapters?} 
We measure \textit{negative transfer} as the increase in 
code perplexity relative
to code-only run.
Below, we demonstrate routing and gradient diagnostics to localize this degradation for Phi-tiny-MoE-instruct on code + biomedical domain pair (corresponding gradient analysis for OLMoE appears in Appendix \ref{app:gradient_cosine}).

\begin{figure*}
    \centering
\includegraphics[width=0.9\linewidth]{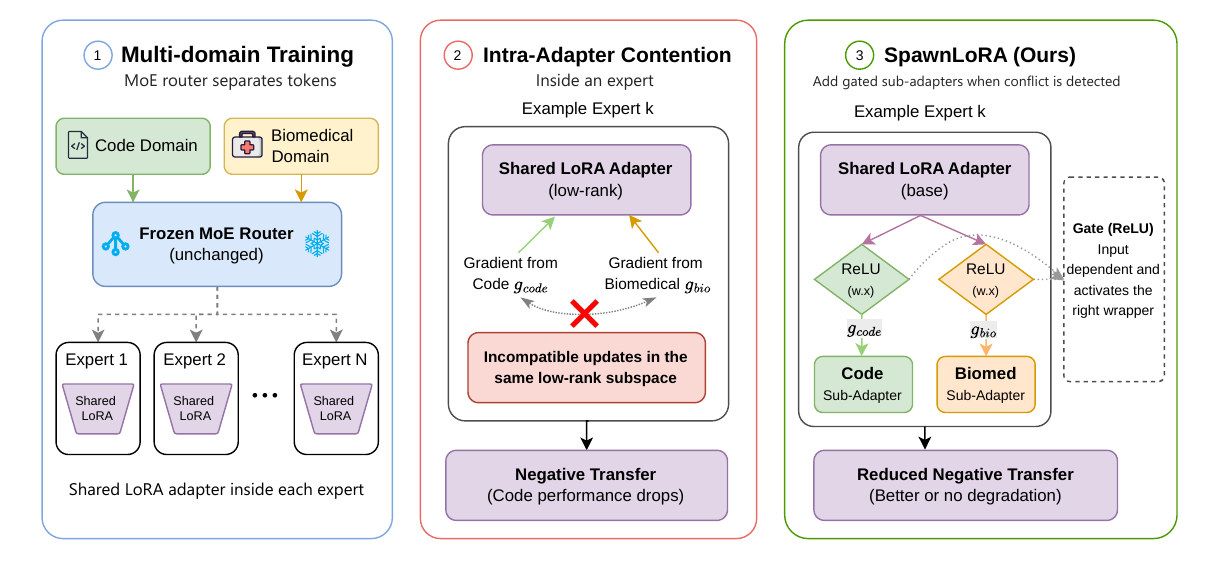}
    \caption{\small{SpawnLoRA overview. Shared LoRA adapters can suffer intra-adapter
contention even when MoE routing separates domains. SpawnLoRA reduces negative
transfer by adding gated sub-adapters inside experts while keeping the router
fixed.}}

    \label{fig:method}
    
\end{figure*}
\paragraph{Ruling out routing.}
We first measure whether the two domains activate the same experts. For layer
$\ell$ and training step $t$, let $S_c^{(\ell,t)}$ and $S_m^{(\ell,t)}$ denote the sets of experts activated by code and biomedical tokens. 
The Jaccard routing overlap value of 0 indicates disjoint routing, while 1 indicates identical expert
usage.
On Phi-tiny-MoE-instruct, the average overlap across layers and training steps is
$\bar{J}\approx0.056$. Thus, the router largely separates the two domains.
Routing failure is therefore an unlikely explanation, and we turn to the adapter updates directly.

\paragraph{Gradient-Space Diagnosis.}
We next test whether the two domains produce compatible updates inside the shared
LoRA adapters. Let $g_c$ and $g_m$ denote the per-domain adapter gradients for
code and biomedical examples. We measure their alignment using cosine similarity, where 
a value of $+1$ indicates aligned updates, $-1$ indicates directly opposed
updates, and 0 indicates orthogonal updates.
Across 20 trials, the aggregate cosine is $\cos(g_c,g_m)=-0.002\pm0.003$, (75\% below zero), showing the domains are unrelated rather than aligned. The domains interfere not because they share routing paths, but because their adapter gradients are nearly orthogonal inside the shared LoRA subspace, a pattern that routing separation cannot prevent.

\section{SpawnLoRA}
\label{sec:spawnlora}
The diagnostic analysis localizes interference inside the shared LoRA adapter. To address this, we propose SpawnLoRA (Fig.~\ref{fig:method}), which adds gated sub-adapters inside an expert when contention is detected. 
\subsection{Method}
\label{sec:spawnlora-method}
\paragraph{Architecture.}
Each expert $E_k$ contains frozen expert weights $W_k$ and an
always-active base LoRA adapter $L_{k,0}$. During training, SpawnLoRA adds gated sub-adapters $L_{k,j}$ inside the same expert. The expert output is:

{\scriptsize
\begin{equation}
E_k(x) = W_k x + L_{k,0}(x) + \sum_{j=1}^{J_k} \mathrm{ReLU}(w_{k,j}^{\top}x) L_{k,j}(x),
\end{equation}
}
where $x$ is the expert input, $L_{k,0}$ is the base LoRA update, and each
$L_{k,j}$ is a separate pathway gated by the learned $\mathrm{ReLU}(w_{k,j}^{\top}x)$ that controls how strongly
the $j$-th sub-adapter contributes for input $x$, enabling input-dependent specialization without changing token routing.

{
\color{black}
\paragraph{Overhead.}
A standard LoRA adapter has parameter and per-token computational cost $O(r\,d_{\mathrm{model}})$. For an expert with $J_k$ spawned sub-adapters, the corresponding cost becomes $O((1+J_k)\,r\,d_{\mathrm{model}})$. Since spawning occurs only in experts where contention is detected and $J_k$ is capped at $J_{\max}=10$, the additional cost remains localized rather than being applied uniformly across all experts. SpawnLoRA also keeps the pretrained router unchanged and does not activate additional experts.
}

\paragraph{Spawn trigger.}
SpawnLoRA adds a sub-adapter only when \textit{two conditions} persist over a rolling window, i.e., the most recent $N$ training steps used to avoid reacting to a single noisy batch. First, the current adapter's importance score stops improving, indicating the current pathway has stopped changing. Second, the gap between the two domains' smoothed losses exceeds $\delta$, indicating the conflict between domains remains unresolved. After spawning, the window resets, and each expert is capped at a maximum number of sub-adapters. Hyperparameters and importance score are provided in Appendix~\ref{app:importance}.
\textcolor{black}{We measured the code/interfering-domain gradient cosine when each spawn occurred. Across $N=20$ spawn events, values ranged from $-0.013$ to $+0.140$ ($82^\circ$--$91^\circ$), showing that the gradients were near-orthogonal rather than aligned.}

\paragraph{Initialization and routing.} Each spawned sub-adapter is a standard LoRA module $L_{k,j}(x)=B_{k,j}A_{k,j}x$. We initialize $A_{k,j}$ from the leading residual-gradient directions not captured by the existing adapter subspace, and set $B_{k,j}=0$ such that the new pathway does not immediately change the model output.
SpawnLoRA adds sub-adapters inside selected experts without changing which tokens are routed to which expert.
\subsection{Results and Discussion}
\begin{table}[t]
\centering
\scriptsize
\setlength{\tabcolsep}{3pt}
\begin{tabular}{llccc}
\toprule
Method & Run & Mix & PPL$_\mathrm{code}$ & NegTransfer \\
\midrule
LoRA      & A & 0\%  & $391.55 \pm 46.04$   & --- \\
DR-LoRA   & A & 0\%  & $505.86 \pm 52.58$\begingroup\renewcommand{\thefootnote}{$\dagger$}\footnotemark\endgroup   & --- \\
SpawnLoRA & A & 0\%  & $425.79 \pm 29.72$   & --- \\
\midrule
LoRA      & B & 20\% & $613.05 \pm 87.16$   & $+221.49 \pm 87.16$ \\
DR-LoRA   & B & 20\% & $837.28 \pm 82.00$   & $+331.41 \pm 54.59$ \\
SpawnLoRA & B & 20\% & $\mathbf{618.03 \pm 23.42}$ & $\mathbf{+192.24 \pm 23.42}$ \\
\midrule
LoRA      & C & 50\% & $1074.72 \pm 132.20$ & $+683.16 \pm 132.20$ \\
DR-LoRA   & C & 50\% & $1633.95 \pm 104.17$ & $+1128.09 \pm 153.82$ \\
SpawnLoRA & C & 50\% & $\mathbf{1039.73 \pm 32.43}$ & $\mathbf{+613.94 \pm 32.43}$ \\
\bottomrule
\end{tabular}
{

\caption{\small{\label{tab:main} Code perplexity
and negative transfer on the code and biomedical text domain pair
for Phi-tiny-MoE-instruct. Lower is better.}}}

\end{table}

\begin{table}[t]
\centering
\scriptsize
\setlength{\tabcolsep}{3pt}
\begin{tabular}{llccccc}
\toprule
Method & Run & Mix & Code PPL & NegTransfer & Expansions \\
\midrule
LoRA      & A & 0\%  & 50.49  & ---              & 0  \\
DR-LoRA   & A & 0\%  & 58.78  & ---              & 6  \\
SpawnLoRA & A & 0\%  & \textbf{47.49} & ---     & 0  \\
\midrule
LoRA      & B & 20\% & 61.17  & $+10.68$         & 0  \\
DR-LoRA   & B & 20\% & 67.12  & $+8.34$         & 6  \\
SpawnLoRA & B & 20\% & \textbf{42.64} & $\mathbf{-4.85}$ & \textcolor{black}{10} \\
\midrule
LoRA      & C & 50\% & 110.14 & $+59.65$         & 0  \\
DR-LoRA   & C & 50\% & 125.39 & $+66.61$         & 6  \\
SpawnLoRA & C & 50\% & \textbf{80.48}  & $\mathbf{+32.99}$ & \textcolor{black}{10} \\
\bottomrule
\end{tabular}
\caption{\small{Code perplexity and negative
transfer on the code and biomedical text domain pair for
OLMoE-1B-7B. Bold indicates best result per conflict level.
NegTransfer is relative to Run~A of the same method; positive
values indicate degradation over no-conflict training. Lower is better.}}
\label{tab:olmoe}
\end{table}

\begin{table}[t]
\centering
\scriptsize
\setlength{\tabcolsep}{3pt}
\begin{tabular}{llccc}
\toprule
Method & Run & Mix & PPL$_\mathrm{code}$ & NegTransfer \\
\midrule
LoRA      & A & 0\%  & 28.33 & --- \\
DR-LoRA   & A & 0\%  & 28.56 & --- \\
SpawnLoRA & A & 0\%  & \textbf{28.44} & --- \\
\midrule
LoRA      & B & 20\% & 31.96 & $+3.63$ \\
DR-LoRA   & B & 20\% & 31.79 & $+3.23$ \\
SpawnLoRA & B & 20\% & \textbf{29.47} & $\mathbf{+1.03}$ \\
\midrule
LoRA      & C & 50\% & 40.80 & $+12.47$ \\
DR-LoRA   & C & 50\% & 39.36 & $+10.80$ \\
SpawnLoRA & C & 50\% & \textbf{36.26} & $\mathbf{+7.82}$ \\
\bottomrule
\end{tabular}
\caption{\small{Code perplexity and negative transfer on the code and mathematical
reasoning (GSM8K) domain pair for OLMoE-1B-7B. Lower is better.
\textcolor{black}{NegTransfer here includes reduced code exposure, not just interference; see
Section~\ref{sec:main-results} for the breakdown.}}}
\label{tab:gsm8k}
\end{table}
\noindent \textbf{Main Results.}\label{sec:main-results}
We compare SpawnLoRA with standard LoRA and the rank adaptive DR-LoRA baseline. Results are reported as the mean $\pm$ standard deviation over three random seeds. For Phi-tiny-MoE-instruct, Table~\ref{tab:main} shows that biomedical text increases code perplexity for both LoRA and DR-LoRA. SpawnLoRA yields the lowest negative transfer in both mixture settings, while DR-LoRA performs worse than fixed-rank LoRA, consistent with our diagnosis that under near-orthogonal gradients, expanding shared rank gives both domains more directions to compete over, which amplifies contention rather than resolving it. \textcolor{black}{This effect compounds with both domain severity and architecture: on OLMoE, DR-LoRA's Run C negative transfer is 6.2$\times$ larger for the more severe biomedical pair than for the milder GSM8K pair ($+66.61$ vs.\ $+10.80$, Tables~\ref{tab:olmoe}--\ref{tab:gsm8k}), and for the biomedical pair specifically, Phi's rank expansion degrades roughly 16.9$\times$ more than OLMoE's ($+1128.09$ vs.\ $+66.61$, Tables~\ref{tab:main}--\ref{tab:olmoe}).} SpawnLoRA instead adds separated gated pathways inside experts, reducing both degradation and seed variance (per-seed results in Appendix~\ref{app:per_seed}) and suggesting that structural separation stabilizes training dynamics in addition to reducing negative transfer.

On OLMoE-1B-7B (Table~\ref{tab:olmoe}) as well, SpawnLoRA eliminates negative transfer in Run B and reduces code perplexity by 26.9\% in Run C. \textcolor{black}{Parameter scaling alone does not explain these gains, as SpawnLoRA's peak capacity for OLMoE equates to a fixed-rank LoRA of only $r \approx 17$ against a base rank of 16 ($\Delta r = 1$; Appendix~\ref{app:parameter_count}). A direct fixed-rank baseline on the code+GSM8K pair (single seed) confirms this, as matching SpawnLoRA's peak parameter count ($r=17$) reduces standard LoRA's own negative transfer by only 6\% (Run B) and 2\% (Run C) relative to the base rank ($r=16$), while SpawnLoRA removes 83\% and 44\% of that same degradation. The equivalent comparison for the Phi-tiny-MoE-instruct results in Table~\ref{tab:main} ($r \approx 21$) remains future work.} With GSM8K as the second domain (Table~\ref{tab:gsm8k}), SpawnLoRA again performs best, reducing code perplexity over LoRA by 7.8\% in Run~B and 11.1\% in Run~C, confirming that intra-adapter contention spans architectures and is not limited to biomedicine. \textcolor{black}{A code-only control matched to Runs B and C in coverage and update count (Section~\ref{sec:setup}) helps separate the effect of reduced code exposure from interference. For LoRA, the control gives 30.29 PPL in Run B and 35.44 PPL in Run C, compared with 32.07 and 40.54 PPL for the corresponding mixed runs. Relative to the code-only reference of 28.51 PPL, this leaves an additional degradation of $+1.78$ PPL in Run B and $+5.10$ PPL in Run C, which is consistent with interference. We use this control only to estimate the possible effect of reduced code exposure and do not subtract it from the reported results. Negative transfer remains positive and non-trivial after accounting for this effect.}

\begingroup
\renewcommand{\thefootnote}{$\dagger$}
\interfootnotelinepenalty=10000
\footnotetext{DR-LoRA's high Run~A perplexity is an artifact of its rank-penalty schedule under batch size~1 (Appendix~\ref{app:drlora})}
\endgroup

\noindent \textbf{Layer-wise expansion as an improvement strategy.}
Table~\ref{tab:multilayer} shows that enabling spawning at additional high-overlap
layers yields further gains on Phi-tiny-MoE-instruct: adding L1 or L3 effectively reduces code perplexity from 684.52 to
361.24 for L1 and 352.45 for L3. Therefore, targeted layer-wise expansion is a practical
strategy when additional compute is available.

\begin{table}[t]
\centering
\small
\begin{tabular}{ccc}
\toprule
PPL$_\mathrm{code}$ & NegTransfer & Active Layers \\
\midrule
{684.52}  & ${+219.60}$ & L0 \\
361.24 & $-323.30$ & L0, L1 \\
352.45 & $-332.09$ & L0, L3 \\
\bottomrule
\end{tabular}
\caption{\small{Multi-layer SpawnLoRA on Phi-tiny-MoE-instruct (Run B). Lower is better.}}
\label{tab:multilayer}
\end{table}

{\color{black}
\noindent\textbf{MBPP and syntactic pass evaluation.}
To verify that the signal is not a perplexity artifact, we report
MBPP token accuracy \citep{austin2021program} and syntactic pass rate,
defined as the fraction of generated outputs that can be parsed successfully
as Python code. These metrics provide useful signal at our small training
scale, where pass@1 remains near zero for all methods.
} 
Under severe conflict on OLMoE, SpawnLoRA recovers MBPP token accuracy to 33.18\% vs. 19.71\% for LoRA  ($+$13.46pp). Syntactic pass shows a similar pattern (Table~\ref{tab:syntactic_pass}): all adapted models degrade relative to the base model, but SpawnLoRA retains the highest rate (30.0\% vs. 22.0\% for LoRA, 2.7\% for DR-LoRA), supporting the perplexity findings.
\begin{table}[t]
\centering
\scriptsize
\setlength{\tabcolsep}{4pt}
\begin{tabular}{lcc}
\toprule
Method & Syntactic pass rate & Drop from base \\
\midrule
Base       & 88.0\% & --- \\
LoRA       & 22.0\% & 66.0 points \\
DR-LoRA    & 2.7\%  & 85.3 points \\
SpawnLoRA  & \textbf{30.0\%} & \textbf{58.0 points} \\
\bottomrule
\end{tabular}
\caption{\small{Syntactic pass rate under the severe conflict setting. Higher is better. SpawnLoRA shows the smallest degradation among the adaptation methods.}}
\label{tab:syntactic_pass}
\end{table}

\section{Conclusion}
\label{sec:conclusion}
We show that successful MoE routing does not prevent multi-domain interference: near-orthogonal domain gradients can still compete within a shared low-rank adapter. SpawnLoRA instead adds gated sub-adapters inside experts, reducing negative transfer relative to fixed- and rank-adaptive LoRA across both models. Multi-layer spawning further eliminates negative transfer under moderate conflict on Phi-tiny-MoE-instruct. These results favour structural separation over shared-capacity expansion. 

\section*{Limitations}
\label{sec:limitations}

Our experiments are designed as a controlled diagnostic study, and the scope below reflects that design. The diagnosis of intra-adapter contention is supported directly by routing and gradient measurements, while the empirical evaluation of SpawnLoRA covers the settings studied here.

\paragraph{Domain coverage.}Primary experiments use two domain pairs: Python code paired with biomedical text and with mathematical reasoning, chosen to create a strong interference signal. This setting is useful for diagnosis, but it does not show whether the same gradient-orthogonality pattern appears in less dissimilar pairings, such as code and formal reasoning or English and Non-English. The diagnostic framework itself is not tied to these domains, and applying it to broader domain pairs is an important next step. Future work should also examine whether the gradient-orthogonality pattern scales with the number of domains, as contention may compound non-linearly when more than two domains compete within the same adapter subspace.

\paragraph{Scale.} Due to computational constraints, our experiments use models in the 1--8B parameter range, train for 3,000 steps on fixed 2,000-example mixtures, and report main results as mean $\pm$ standard deviation over three random seeds $\{42, 123, 456\}$. Whether the diagnosis and SpawnLoRA's mitigation remain stable at larger scales, such as $\geq$70B parameters or substantially longer training runs, and whether the spawn trigger requires recalibration in such regimes, remains open.

\paragraph{Functional evaluation.} Our headline results use perplexity. Pass@1 evaluation on HumanEval was not informative at this training scale, since strong code fine-tuning typically requires tens of thousands of examples, while our controlled interference experiments use fixed 2,000-example mixtures. Downstream task accuracy on PubMedQA was similarly uninformative at this training scale, consistent with the limited number of biomedical training examples used. We therefore include syntactic pass and MBPP as additional functional checks for code, while treating perplexity as the metric most directly aligned with our core claim that structural separation reduces cross-domain interference. These functional evaluations are intended to complement the main analysis, and broader downstream evaluation remains future work for settings with sufficient per-domain data.

\paragraph{Architectural coverage.} Due to computational constraints, our evaluation is limited to two MoE families. Generalisation to architectures with very different routing granularity, such as DeepSeek-V2~\citep{deepseek2024deepseekv2}, remains open. Our cross-architecture preliminary analysis suggests that routing granularity affects intra-adapter conflict, but expanding this analysis to more architectures remains an important extension of this work. In particular, architectures with shared-expert designs, such as DeepSeek-V2, may exhibit qualitatively different contention patterns because some experts receive tokens from all domains by construction, making structural separation inside those experts both more necessary and more complex to implement.

\section*{Ethical Considerations}

This work studies parameter efficient fine tuning for pretrained language models in a controlled research setting. The experiments use publicly available datasets and open model checkpoints to examine cross domain interference in MoE and LoRA fine tuning. The datasets used in our experiments are HumanEval, PubMedQA, and GSM8K. HumanEval, PubMedQA, and GSM8K are distributed under MIT licenses. These resources are used only for research and in accordance with their stated terms. We do not collect new human subject data, conduct user studies, or process personally identifiable information. PubMedQA is used only as biomedical text for controlled experimentation and should not be interpreted as clinical validation, medical advice, or evidence of suitability for clinical use.

The primary societal risk of this work is that more efficient adaptation methods may support the specialization of language models for sensitive domains, including biomedical text processing and code generation. Models adapted with these methods may inherit factual errors, biases, unsafe code patterns, or other limitations from their underlying models and data. We did not evaluate these failure modes directly, and the adapted models are not intended for deployment in medical, safety critical software, legal, or other high stakes settings without independent safety, robustness, fairness, privacy, and domain specific evaluation.
All experiments use publicly available components and are intended to be reproducible from the reported setup. Phi Tiny MoE training and evaluation were conducted on a single NVIDIA L4 GPU, with observed peak memory use of 13.9 GB. OLMoE training and evaluation were conducted on a single NVIDIA A100 GPU, with observed peak memory use of 27.8 GB. Code will be released upon acceptance. The manuscript is written for human readers and contains no hidden instructions intended to influence automated reviewing or processing systems.

\section*{Acknowledgments}

Our work was supported by the Islamic University of Technology Research Seed Grants (Ref: REASP/IUT-RSG/2022/OL/07/012).

We used ChatGPT and Claude to assist with grammar checking, sentence rewording, and language refinement of text originally written by the authors. These tools were not used to generate the paper’s main ideas, experimental design, analysis, or conclusions. ChatGPT was also used for limited assistance with code development. All AI assisted text and code were reviewed, edited, and validated by the authors, who take full responsibility for the final content of the paper.
\bibliography{custom}

\begin{thebibliography}{20}
\providecommand{\natexlab}[1]{#1}

\bibitem[{Austin et~al.(2021)Austin, Odena, Nye, Bosma, Michalewski, Dohan,
  Jiang, Cai, Terry, Le, and Sutton}]{austin2021program}
Jacob Austin, Augustus Odena, Maxwell Nye, Maarten Bosma, Henryk Michalewski,
  David Dohan, Ellen Jiang, Carrie Cai, Michael Terry, Quoc Le, and Charles
  Sutton. 2021.
\newblock \href {https://arxiv.org/abs/2108.07732} {Program synthesis with
  large language models}.
\newblock \emph{arXiv preprint arXiv:2108.07732}.

\bibitem[{Chen et~al.(2021)Chen, Tworek, Jun, Yuan, Ponde~de Oliveira~Pinto,
  Kaplan, Edwards, Burda, Joseph, Brockman, Ray, Puri, Krueger, Petrov, Khlaaf,
  Sastry, Mishkin, Chan, Gray, Ryder, Pavlov, Power, Kaiser, Bavarian, Winter,
  Tillet, Such, Cummings, Plappert, Chantzis, Barnes, Herbert-Voss, Guss,
  Nichol, Paino, Tezak, Tang, Babuschkin, Balaji, Jain, Saunders, Hesse, Carr,
  Leike, Achiam, Misra, Morikawa, Radford, Knight, Brundage, Murati, Mayer,
  Welinder, McGrew, Amodei, McCandlish, Sutskever, and
  Zaremba}]{chen2021evaluatinglargelanguagemodels}
Mark Chen, Jerry Tworek, Heewoo Jun, Qiming Yuan, Henrique Ponde~de
  Oliveira~Pinto, Jared Kaplan, Harri Edwards, Yuri Burda, Nicholas Joseph,
  Greg Brockman, Alex Ray, Raul Puri, Gretchen Krueger, Michael Petrov, Heidy
  Khlaaf, Girish Sastry, Pamela Mishkin, Brooke Chan, Scott Gray, and 39
  others. 2021.
\newblock \href {https://arxiv.org/abs/2107.03374} {Evaluating large language
  models trained on code}.
\newblock \emph{arXiv preprint arXiv:2107.03374}.

\bibitem[{Cobbe et~al.(2021)Cobbe, Kosaraju, Bavarian, Chen, Jun, Kaiser,
  Plappert, Tworek, Hilton, Nakano, Hesse, and
  Schulman}]{cobbe2021trainingverifierssolvemath}
Karl Cobbe, Vineet Kosaraju, Mohammad Bavarian, Mark Chen, Heewoo Jun, Lukasz
  Kaiser, Matthias Plappert, Jerry Tworek, Jacob Hilton, Reiichiro Nakano,
  Christopher Hesse, and John Schulman. 2021.
\newblock \href {https://arxiv.org/abs/2110.14168} {Training verifiers to solve
  math word problems}.
\newblock \emph{arXiv preprint arXiv:2110.14168}.

\bibitem[{{DeepSeek-AI}(2024)}]{deepseek2024deepseekv2}
{DeepSeek-AI}. 2024.
\newblock \href {https://arxiv.org/abs/2405.04434} {{DeepSeek-V2}: A strong,
  economical, and efficient mixture-of-experts language model}.
\newblock \emph{arXiv preprint arXiv:2405.04434}.

\bibitem[{Deng et~al.(2026)Deng, Li, Chen, Wang, Song, and Wen}]{liu2024drlora}
Guanzhi Deng, Bo~Li, Ronghao Chen, Huacan Wang, Linqi Song, and Lijie Wen.
  2026.
\newblock \href {https://arxiv.org/abs/2601.04823} {{DR-LoRA}: Dynamic rank
  {LoRA} for mixture-of-experts adaptation}.
\newblock \emph{arXiv preprint arXiv:2601.04823}.

\bibitem[{Dou et~al.(2024)Dou, Zhou, Liu, Gao, Shen, Xiong, Zhou, Wang, Xi,
  Fan, Pu, Zhu, Zheng, Gui, Zhang, and Huang}]{dou2024loramoe}
Shihan Dou, Enyu Zhou, Yan Liu, Songyang Gao, Wei Shen, Limao Xiong, Yuhao
  Zhou, Xiao Wang, Zhiheng Xi, Xiaoran Fan, Shiliang Pu, Jiang Zhu, Rui Zheng,
  Tao Gui, Qi~Zhang, and Xuanjing Huang. 2024.
\newblock \href {https://doi.org/10.18653/v1/2024.acl-long.106} {{LoRAMoE}:
  Alleviating world knowledge forgetting in large language models via
  {MoE}-style plugin}.
\newblock In \emph{Proceedings of the 62nd Annual Meeting of the Association
  for Computational Linguistics (Volume 1: Long Papers)}, pages 1932--1945,
  Bangkok, Thailand. Association for Computational Linguistics.

\bibitem[{Fedus et~al.(2022)Fedus, Zoph, and Shazeer}]{fedus2022switch}
William Fedus, Barret Zoph, and Noam Shazeer. 2022.
\newblock \href {https://jmlr.org/papers/v23/21-0998.html} {Switch
  transformers: Scaling to trillion parameter models with simple and efficient
  sparsity}.
\newblock \emph{Journal of Machine Learning Research}, 23(120):1--39.

\bibitem[{Han et~al.(2024)Han, Gao, Liu, Zhang, and Zhang}]{hanparameter}
Zeyu Han, Chao Gao, Jinyang Liu, Jeff Zhang, and Sai~Qian Zhang. 2024.
\newblock Parameter-efficient fine-tuning for large models: A comprehensive
  survey.
\newblock \emph{Transactions on Machine Learning Research}.

\bibitem[{Hu et~al.(2022)Hu, Shen, Wallis, Allen-Zhu, Li, Wang, Wang, and
  Chen}]{hu2022lora}
Edward~J. Hu, Yelong Shen, Phillip Wallis, Zeyuan Allen-Zhu, Yuanzhi Li, Shean
  Wang, Lu~Wang, and Weizhu Chen. 2022.
\newblock \href {https://openreview.net/forum?id=nZeVKeeFYf9} {{LoRA}: Low-rank
  adaptation of large language models}.
\newblock In \emph{International Conference on Learning Representations}.

\bibitem[{Huynh et~al.(2025)Huynh, Vu, Wang, Le, Gasevic, Li, and
  Do}]{huynh2025mixlora}
Tuan-Luc Huynh, Thuy-Trang Vu, Weiqing Wang, Trung Le, Dragan Gasevic,
  Yuan-Fang Li, and Thanh-Toan Do. 2025.
\newblock \href {https://doi.org/10.18653/v1/2025.emnlp-main.20}
  {{MixLoRA}-{DSI}: Dynamically expandable mixture-of-{LoRA} experts for
  rehearsal-free generative retrieval over dynamic corpora}.
\newblock In \emph{Proceedings of the 2025 Conference on Empirical Methods in
  Natural Language Processing}, pages 380--396, Suzhou, China. Association for
  Computational Linguistics.

\bibitem[{Jin et~al.(2019)Jin, Dhingra, Liu, Cohen, and
  Lu}]{jin2019pubmedqadatasetbiomedicalresearch}
Qiao Jin, Bhuwan Dhingra, Zhengping Liu, William Cohen, and Xinghua Lu. 2019.
\newblock \href {https://doi.org/10.18653/v1/D19-1259} {{PubMedQA}: A dataset
  for biomedical research question answering}.
\newblock In \emph{Proceedings of the 2019 Conference on Empirical Methods in
  Natural Language Processing and the 9th International Joint Conference on
  Natural Language Processing (EMNLP-IJCNLP)}, pages 2567--2577, Hong Kong,
  China. Association for Computational Linguistics.

\bibitem[{Li et~al.(2024)Li, Ma, Wang, Ye, Cheng, Tang, Zhang, Duan, Zuo, Yang,
  and Tang}]{zhu2024mixlora}
Dengchun Li, Yingzi Ma, Naizheng Wang, Zhengmao Ye, Zhiyuan Cheng, Yinghao
  Tang, Yan Zhang, Lei Duan, Jie Zuo, Cal Yang, and Mingjie Tang. 2024.
\newblock \href {https://arxiv.org/abs/2404.15159} {{MixLoRA}: Enhancing large
  language models fine-tuning with {LoRA}-based mixture of experts}.
\newblock \emph{arXiv preprint arXiv:2404.15159}.

\bibitem[{Li et~al.(2025)Li, Liang, Zhang, Hong, Kim, Chen, and
  Zhao}]{lislimmoe}
Zichong Li, Chen Liang, Zixuan Zhang, Ilgee Hong, Young~Jin Kim, Weizhu Chen,
  and Tuo Zhao. 2025.
\newblock {SlimMoE}: Structured compression of large {MoE} models via expert
  slimming and distillation.
\newblock In \emph{Second Conference on Language Modeling}.

\bibitem[{Muennighoff et~al.(2025)Muennighoff, Soldaini, Groeneveld, Lo,
  Morrison, Min, Shi, Walsh, Tafjord, Lambert, Gu, Arora, Bhagia, Schwenk,
  Wadden, Wettig, Hui, Dettmers, Kiela, Farhadi, Smith, Koh, Singh, and
  Hajishirzi}]{muennighoff2024olmoe}
Niklas Muennighoff, Luca Soldaini, Dirk Groeneveld, Kyle Lo, Jacob Morrison,
  Sewon Min, Weijia Shi, Pete Walsh, {\O}yvind Tafjord, Nathan Lambert, Yuling
  Gu, Shane Arora, Akshita Bhagia, Dustin Schwenk, David Wadden, Alexander
  Wettig, Binyuan Hui, Tim Dettmers, Douwe Kiela, and 5 others. 2025.
\newblock {OLMoE}: Open mixture-of-experts language models.
\newblock In \emph{International Conference on Learning Representations}.

\bibitem[{Peng et~al.(2026)Peng, Zou, Zeng, Li, Chen, Li, and
  Wang}]{peng2026hilora}
Zihao Peng, Nan Zou, Jiandian Zeng, Guo Li, Ke~Chen, Boyuan Li, and Tian Wang.
  2026.
\newblock {HiLoRA}: Hierarchical low-rank adaptation for personalized federated
  learning.
\newblock In \emph{Proceedings of the IEEE/CVF Conference on Computer Vision
  and Pattern Recognition}, pages 31746--31757.

\bibitem[{Shazeer et~al.(2017)Shazeer, Mirhoseini, Maziarz, Davis, Le, Hinton,
  and Dean}]{shazeer2017outrageously}
Noam Shazeer, Azalia Mirhoseini, Krzysztof Maziarz, Andy Davis, Quoc~V. Le,
  Geoffrey~E. Hinton, and Jeff Dean. 2017.
\newblock \href {https://openreview.net/forum?id=B1ckMDqlg} {Outrageously large
  neural networks: The sparsely-gated mixture-of-experts layer}.
\newblock In \emph{International Conference on Learning Representations}.

\bibitem[{Wang et~al.(2021)Wang, Tsvetkov, Firat, and Cao}]{wang2021gradient}
Zirui Wang, Yulia Tsvetkov, Orhan Firat, and Yuan Cao. 2021.
\newblock \href {https://openreview.net/forum?id=F1vEjWK-lH_} {Gradient
  vaccine: Investigating and improving multi-task optimization in massively
  multilingual models}.
\newblock In \emph{International Conference on Learning Representations}.

\bibitem[{Yu et~al.(2020)Yu, Kumar, Gupta, Levine, Hausman, and
  Finn}]{yu2020gradient}
Tianhe Yu, Saurabh Kumar, Abhishek Gupta, Sergey Levine, Karol Hausman, and
  Chelsea Finn. 2020.
\newblock \href
  {https://proceedings.neurips.cc/paper/2020/hash/3fe78a8acf5fda99de95303940a2420c-Abstract.html}
  {Gradient surgery for multi-task learning}.
\newblock In \emph{Advances in Neural Information Processing Systems},
  volume~33, pages 5824--5836.

\bibitem[{Zadouri et~al.(2024)Zadouri, {\"U}st{\"u}n, Ahmadian, Ermi{\c{s}},
  Locatelli, and Hooker}]{zadouri2024pushing}
Ted Zadouri, Ahmet {\"U}st{\"u}n, Arash Ahmadian, Beyza Ermi{\c{s}}, Acyr
  Locatelli, and Sara Hooker. 2024.
\newblock \href {https://openreview.net/forum?id=EvDeiLv7qc} {Pushing mixture
  of experts to the limit: Extremely parameter efficient {MoE} for instruction
  tuning}.
\newblock In \emph{International Conference on Learning Representations}.

\bibitem[{Zhang et~al.(2023)Zhang, Chen, Bukharin, He, Cheng, Chen, and
  Zhao}]{zhang2023adalora}
Qingru Zhang, Minshuo Chen, Alexander Bukharin, Pengcheng He, Yu~Cheng, Weizhu
  Chen, and Tuo Zhao. 2023.
\newblock \href {https://openreview.net/forum?id=lq62uWRJjiY} {Adaptive budget
  allocation for parameter-efficient fine-tuning}.
\newblock In \emph{International Conference on Learning Representations}.

\end{thebibliography}

\appendix

\appendix

\section{Reproducibility Details}
\label{app:reproducibility}

\subsection{Hyperparameter Values}
\label{app:hyperparams}

Table~\ref{tab:hyperparams} lists the SpawnLoRA trigger and
initialisation hyperparameters.
\textcolor{black}{The loss-gap threshold~$\delta$ differs by model, while the base rank~$r$
and all other values are shared across architectures.}
Table~\ref{tab:training} lists the training hyperparameters common to
all three methods (LoRA, DR-LoRA, SpawnLoRA) unless noted.

\begin{table}[H]
\centering
\scriptsize
\begin{tabular}{lp{3cm}p{2.7cm}}
\toprule
Symbol & Description & Value \\
\midrule
$r$        & Base LoRA rank
           & 16 \\
\textcolor{black}{$r_s$} & \textcolor{black}{Sub-adapter (spawn) rank}
           & \textcolor{black}{8 (shared)} \\
$W$        & Rolling window for spawn trigger (steps)
           & 15 \\
$\tau$     & Importance-score improvement threshold
           & $10^{-4}$(Phi-tiny-MoE-instruct) / $10^{-3}$(OLMoE) \\
$\beta$    & EMA smoothing factor for per-domain loss gap
           & 0.9 \\
\textcolor{black}{$\delta$} & \textcolor{black}{Per-domain loss threshold (nats)}
           & \textcolor{black}{1.0 (Phi-tiny-MoE-instruct) / 0.3 (OLMoE)} \\
$J_{\max}$ & Maximum sub-adapters per expert
           & 10 \\
\midrule
$\sigma_w$ & Gate initialisation scale for $w_{k,j}$
           & $10^{-3} \cdot \mathrm{Var}(W_k)$ \\
\bottomrule
\end{tabular}
\caption{SpawnLoRA-specific dynamic allocation and initialization hyperparameters. For parameters denoting dual values, the primary value corresponds to the Phi-tiny-MoE-instruct architecture, while the secondary value applies to both OLMoE-1B-7B.}
\label{tab:hyperparams}
\end{table}

\begin{table}[H]
\centering
\scriptsize
\begin{tabular}{lp{1.6cm}p{3.0cm}}
\toprule
Hyperparameter & Value & Notes \\
\midrule
Optimizer      & AdamW & \\
Learning rate  & $2{\times}10^{-5}$ / $5{\times}10^{-5}$
               & Phi-tiny-MoE-instruct / OLMoE \\
Batch size     & 1     & \\
LR schedule    & constant & no warmup or decay \\
LoRA $\alpha$  & $2r$  & 32 (Phi-tiny-MoE-instruct) / 16 (OLMoE) \\
LoRA dropout   & 0.0   & SpawnLoRA and LoRA; DR-LoRA uses 0.1 \\
Target modules & \texttt{down\_proj}
               & SpawnLoRA and LoRA; DR-LoRA targets all MLP
                 projections \\
\bottomrule
\end{tabular}
\caption{Baseline optimization configurations and general training hyperparameters. Unless explicitly specified in the notes, these values remain constant across all evaluated adaptation methods (LoRA, DR-LoRA, and SpawnLoRA) and model architectures.}
\label{tab:training}
\end{table}

\paragraph{Selection notes.}
The window~$W$ and importance threshold~$\tau$ jointly control how
quickly stalled adapters trigger a spawn; results were insensitive to
$W \in [10,25]$ and $\tau \in [10^{-5}, 10^{-3}]$.
After each spawn event the rolling window resets, acting as a
refractory mechanism that prevents consecutive allocations within fewer
than $W$ steps.
The loss-gap threshold~$\delta$ was calibrated on a held-out
validation split; the lower value for OLMoE reflects its
finer-grained routing, which reduces within-expert contention and
warrants a more sensitive trigger.
The gate scale $\sigma_w = 10^{-3}{\cdot}\mathrm{Var}(W_k)$ is
per-expert, computed from the frozen weight matrix~$W_k$, and keeps
gate activations commensurate with the existing weight scale.

\subsection{Importance Scores for Dynamic Capacity Allocation}
\label{app:importance}

We use the term \emph{importance score} to denote a scalar that quantifies how
much a LoRA rank direction currently matters to the model's output, used to
decide when or where to allocate more capacity. DR-LoRA and SpawnLoRA both use
importance scores, but they attach different meanings to this scalar.

\paragraph{DR-LoRA.}
DR-LoRA uses a gradient--parameter sensitivity score to decide where to add rank
within an existing shared adapter. For rank direction~$j$, with corresponding
LoRA factors $A_j$ and $B_j$, the per-rank sensitivity is
\begin{equation}
s_j = \|\nabla_{A_j} \odot A_j\|_1 \, \|\nabla_{B_j} \odot B_j\|_1,
\end{equation}
where $\odot$ denotes element-wise multiplication. This score is large when a
rank direction both affects the loss and has non-trivial parameter magnitude. At
expert~$i$ in layer~$l$, DR-LoRA averages over the active rank directions,
\begin{equation}
g^{(l,i)} = \frac{1}{r} \sum_{j=1}^{r} s_j,
\end{equation}
and smooths the score over time using an exponential moving average,
\begin{equation}
g_t = \beta g_{t-1} + (1-\beta) s_t.
\end{equation}
The resulting expert score is combined with the routing frequency $f^{(l,i)}$ to
form a saliency score,
\begin{equation}
S^{(l,i)} = \frac{f^{(l,i)} g^{(l,i)}}{(r+1)^\gamma}.
\end{equation}
The rank penalty $(r+1)^\gamma$ discourages repeatedly expanding experts that
already have many active ranks. At each growth event, the top-$k$ experts by
saliency receive new rank slots.

\paragraph{SpawnLoRA.}
SpawnLoRA instead uses a parameter-norm importance score to detect whether the
current base adapter has saturated. For expert~$k$, let $A_{k,0}$ and $B_{k,0}$
denote the base adapter matrices. For rank component~$r$, we compute
\begin{equation}
q_{k,r}=\|B_{k,0,:,r}\|_1\,\|A_{k,0,r,:}\|_1,
\end{equation}
and define the expert-level importance score as
\begin{equation}
g_k=\frac{1}{R}\sum_{r=1}^{R} q_{k,r}.
\end{equation}
Unlike DR-LoRA, this score does not use gradients. It measures the output-scale
contribution of each rank to the current low-rank update
$\Delta W_k=B_{k,0}A_{k,0}$. The score feeds into a bivariate spawn trigger
rather than a rank-growth decision. A plateau is detected when the slope of
$g_k$ over the rolling window satisfies
\begin{equation}
|d g_k/dt|<\tau_{\mathrm{plateau}},
\end{equation}
and a spawn is triggered only when this plateau condition is accompanied by a
domain-conflict condition,
\begin{equation}
|\mathrm{EMA}(L_A)-\mathrm{EMA}(L_B)|>\delta,
\end{equation}
Thus SpawnLoRA allocates a new sub-adapter only when the existing adapter has
stopped changing and the gap between the two domains' smoothed losses remains
above the conflict threshold.

\subsection{Parameter Count Analysis}
\label{app:parameter_count}

SpawnLoRA dynamically allocates parameters during training. Each spawned
sub-adapter introduces $2 \cdot r_s \cdot d_{\mathrm{model}} + d_{\mathrm{model}}$
trainable parameters, accounting for the low-rank projection matrices ($A$
and $B$) and the gating vector, where $r_s$ is the spawned sub-adapter rank.

\paragraph{OLMoE-1B-7B.}
\textcolor{black}{With $d_{\mathrm{model}}=2048$, 64 experts per layer, and rank $r=16$, the
base LoRA configuration requires 4,194,304 parameters. On the
code\,+\,biomedical domain pair, the saturation monitor reached the cap of
$J_{\max}=10$ spawns in Runs B and C. Each spawn adds 34,816 parameters, so
the peak trainable count was 4,542,464.}

\paragraph{Phi-tiny-MoE-instruct.}
With $d_{\mathrm{model}}=3072$, 16 experts per layer, and rank $r=16$, the base LoRA configuration also requires 1,572,864 parameters. The saturation monitor reached the cap of $J_{\max}=10$ in both Runs B and C. Each spawn adds 52,224 parameters, so the peak trainable count equalled the theoretical maximum of 2,095,104.

\begin{table}[!h]
\centering
\scriptsize
\begin{tabular}{lp{3.2cm}}
\toprule
Component & OLMoE-1B-7B \\
\midrule
\textcolor{black}{Architecture} & \textcolor{black}{64 experts $\times$ rank 16, $d_{\mathrm{model}}=2048$} \\
\textcolor{black}{Base LoRA} & \textcolor{black}{$2{\times}16{\times}2048{\times}64 = 4{,}194{,}304$} \\
\textcolor{black}{Per spawn} & \textcolor{black}{$2{\times}8{\times}2048 + 2048 = 34{,}816$} \\
\textcolor{black}{Spawns in Runs B\,\&\,C} & \textcolor{black}{10} \\
\textcolor{black}{Peak trainable params} & \textcolor{black}{$4{,}194{,}304 + 10{\times}34{,}816 = 4{,}542{,}464$} \\
\textcolor{black}{Theoretical max ($J_{\max}=10$)} & \textcolor{black}{$4{,}542{,}464$} \\
\bottomrule
\end{tabular}
\caption{OLMoE-1B-7B trainable parameter count breakdown.
\textcolor{black}{OLMoE reaches the cap of $J_{\max}=10$ in Runs B and C,
matching Phi-tiny-MoE-instruct (Table~\ref{tab:parameter_count_phi}).}}
\label{tab:parameter_count_olmoe}
\end{table}

\begin{table}[!h]
\centering
\scriptsize
\begin{tabular}{lp{3.2cm}}
\toprule
Component & Phi-tiny-MoE-instruct \\
\midrule
Architecture & 16 experts, $d_{\mathrm{model}}=3072$ \\
Base LoRA (single layer) & $2{\times}16{\times}3072{\times}16 = 1{,}572{,}864$ \\
Per spawn ($r_s=8$) & $2{\times}8{\times}3072 + 3072 = 52{,}224$ \\
Spawns in Runs B\,\&\,C & 10 (cap reached) \\
Peak trainable params & $1{,}572{,}864 + 10{\times}52{,}224 = 2{,}095{,}104$ \\
Theoretical max ($J_{\max}=10$) & $2{,}095{,}104$ \\
\bottomrule
\end{tabular}
\caption{Phi-tiny-MoE-instruct trainable parameter count breakdown for the single
patched layer. \textcolor{black}{Phi reaches the cap under both moderate and severe conflict,
suggesting $J_{\max}$ is a binding constraint under sustained conflict.}}
\label{tab:parameter_count_phi}
\end{table}

\paragraph{Equivalent fixed-rank baseline.}
To address the question of whether SpawnLoRA's gains are attributable
to increased parameter count rather than structural separation, we
compute the fixed rank~$r$ that would yield the same total parameter
count as SpawnLoRA at peak allocation, assuming uniform rank across
all experts in the patched layer.

For Phi-tiny-MoE-instruct, SpawnLoRA peaks at 2,095,104 parameters
(cap of $J_{\max}=10$ spawns in a single expert). A fixed-rank LoRA
matching this budget would require rank
$r_{\mathrm{equiv}} = 2{,}095{,}104\,/\,(2 \times 3072 \times 16) \approx 21$.
\textcolor{black}{For OLMoE-1B-7B, the equivalent is
$r_{\mathrm{equiv}} = 4{,}542{,}464\,/\,(2 \times 2048 \times 64) \approx 17$,
one above the base rank of~16. ($r=17$ corresponds to 4,456,448 parameters, 1.9\%
below this true peak, since 17.33 rounds down.)}

Two structural differences favour SpawnLoRA in this comparison: its extra
parameters are concentrated in the single conflicted expert rather than
distributed uniformly, and the OLMoE result shows meaningful gains with
only $\Delta r = 1$ overhead, suggesting the gains are not parameter-driven.
\textcolor{black}{We confirm this directly for OLMoE with a fixed-rank baseline
(Section~\ref{sec:main-results}). A direct fixed-$r=21$ comparison for
Phi-tiny-MoE-instruct remains future work.}

\subsection{Dataset Provenance}
\label{app:datasets}

Table~\ref{tab:datasets} lists the datasets used across all experiments.
All datasets are publicly available under permissive licences and were
accessed via the Hugging Face Datasets library without modification.

\begin{table}[ht]
\centering
\scriptsize
\begingroup
\setlength{\tabcolsep}{3pt}
\begin{tabular*}{\columnwidth}{@{\extracolsep{\fill}}p{1.7cm}p{1.8cm}p{1.4cm}p{2.1cm}@{}}
\toprule
Dataset & Split used & Licence & Role \\
\midrule
HumanEval~\citep{chen2021evaluatinglargelanguagemodels}
  & \texttt{test} (164) & MIT & \hspace{0pt}Primary domain: Python code \\
PubMedQA~\citep{jin2019pubmedqadatasetbiomedicalresearch}
  & \texttt{pqa\_\allowbreak labeled/\allowbreak train} & MIT & \hspace{0pt}Conflict domain: bio\-medical text \\
GSM8K~\citep{cobbe2021trainingverifierssolvemath}
  & \texttt{train} & MIT & \hspace{0pt}Conflict domain: math reasoning \\
\bottomrule
\end{tabular*}
\endgroup
\caption{Datasets used in SpawnLoRA experiments. Sample counts refer to the
number of examples drawn per training run (up to 1,000 per domain). No
personally identifiable information is present in any dataset.}
\label{tab:datasets}
\end{table}
\paragraph{Compute environment.}
All experiments were conducted on single-GPU instances.
Phi-tiny-MoE-instruct experiments ran on a single NVIDIA L4 GPU
with observed peak VRAM of 13.9\,GB.
OLMoE-1B-7B experiments ran on a single NVIDIA A100 GPU
with observed peak VRAM of 27.8\,GB.
No distributed training framework such as DeepSpeed or FSDP was used.

\subsection{Per-Seed Results for Phi-tiny-MoE-instruct}
\label{app:per_seed}

\begin{table}[h]
\centering
\tiny
\setlength{\tabcolsep}{2.5pt}
\begin{tabular}{llcccc}
\toprule
Method & Run & S42 & S123 & S456 & Mean $\pm$ Std \\
\midrule
LoRA      & A & 430.60 & 418.14 & 325.92 & $391.55 \pm 46.04$ \\
LoRA      & B & 747.85 & 527.60 & 563.70 & $613.05 \pm 87.16$ \\
LoRA      & C & 1209.38 & 913.12 & 1101.65 & $1074.72 \pm 132.20$ \\
\midrule
DR-LoRA   & A & 461.10 & 583.50 & 472.99 & $505.86 \pm 52.58$ \\
DR-LoRA   & B & 723.61 & 919.87 & 868.36 & $837.28 \pm 82.00$ \\
DR-LoRA   & C & 1773.03 & 1524.01 & 1604.81 & $1633.95 \pm 104.17$ \\
\midrule
SpawnLoRA & A & 464.92 & 420.42 & 392.04 & $425.79 \pm 29.72$ \\
SpawnLoRA & B & 684.52 & 583.02 & 586.54 & $618.03 \pm 23.42$ \\
SpawnLoRA & C & 1039.72 & 1038.77 & 1040.70 & $1039.73 \pm 32.43$ \\
\bottomrule
\end{tabular}
\caption{Per-seed code perplexity (PPL$_\mathrm{code}$) on Phi-tiny-MoE-instruct
(code + biomedical text domain pair). Lower is better. Mean and standard
deviation are computed over seeds $\{42, 123, 456\}$ and match Table~1 of
the main paper.}
\label{tab:seed_ppl}
\end{table}

\begin{table}[h]
\centering
\tiny
\setlength{\tabcolsep}{2.5pt}
\begin{tabular}{llcccc}
\toprule
Method & Run & S42 & S123 & S456 & Mean $\pm$ Std \\
\midrule
LoRA      & B & $+317.25$ & $+109.46$ & $+237.77$ & $+221.49 \pm 87.16$ \\
LoRA      & C & $+778.78$ & $+494.98$ & $+775.72$ & $+683.16 \pm 132.20$ \\
\midrule
DR-LoRA   & B & $+262.51$ & $+336.37$ & $+395.36$ & $+331.41 \pm 54.59$ \\
DR-LoRA   & C & $+1311.93$ & $+940.52$ & $+1131.81$ & $+1128.09 \pm 153.82$ \\
\midrule
SpawnLoRA & B & $+219.60$ & $+162.60$ & $+194.51$ & $+192.24 \pm 23.42$ \\
SpawnLoRA & C & $+574.79$ & $+618.35$ & $+648.67$ & $+613.94 \pm 32.43$ \\
\bottomrule
\end{tabular}
\caption{Per-seed negative transfer on Phi-tiny-MoE-instruct.
Positive values indicate degradation vs.\ Run~A.
SpawnLoRA variance is lowest (Std~$\leq 32.43$) vs.\
$87.16$ (LoRA) and $153.82$ (DR-LoRA) under severe conflict.}
\label{tab:seed_nt}
\end{table}

Tables~\ref{tab:seed_ppl}--\ref{tab:seed_nt} report mean~$\pm$ std over seeds $\{42, 123, 456\}$
for all three methods on Phi-tiny-MoE-instruct (code + biomedical domain pair).
Negative transfer (NT) is $\mathrm{PPL}_{\mathrm{code}}(\text{Run~X}) - \mathrm{PPL}_{\mathrm{code}}(\text{Run~A})$.

\clearpage
\onecolumn
\section{Algorithm}
\label{app:algorithm}

%
\subsection{SpawnLoRA Algorithm}
\label{app:pseudocode}

\begingroup
\small
\captionof{algorithm}{SpawnLoRA}
\label{alg:spawnlora}
\begin{algorithmic}[1]

\Statex \textbf{Initialisation}
\For{each MoE expert $k$}
  \State Freeze $W_k$
         \Comment{original expert weights}
  \State $A_{k,0} \sim \mathrm{Kaiming\_uniform}(\mathrm{rank},\,d_{\mathrm{in}})$;\quad
         $B_{k,0} \leftarrow \mathbf{0}_{d_{\mathrm{out}}\times\mathrm{rank}}$
         \Comment{$B=0$ ensures zero contribution at $t=0$}
  \State $\texttt{spawn\_loras}[k] \leftarrow [\,]$;\quad
         $\texttt{spawn\_gates}[k] \leftarrow [\,]$
  \State $\texttt{monitor}[k] \leftarrow \mathrm{SaturationMonitor}(W,\tau_{\mathrm{plateau}},\delta)$
\EndFor

\Statex \textbf{Forward pass} \textsc{ExpertForward}$_k(x)$
\State $\mathrm{out} \leftarrow W_k x$
       \Comment{frozen base}
\State $\mathrm{out} \mathrel{+}= (B_{k,0}A_{k,0}x)\cdot\mathrm{scaling}$
       \Comment{base LoRA}
\For{each spawned sub-adapter $j \in \texttt{spawn\_loras}[k]$}
  \State $\mathrm{gate}_j \leftarrow \mathrm{ReLU}(w_{k,j}^{\!\top}x)$
         \Comment{scalar per token}
  \State $\mathrm{out} \mathrel{+}= \mathrm{gate}_j\cdot(B_{k,j}A_{k,j}x)\cdot\mathrm{scaling}$
\EndFor
\State \Return $\mathrm{out}$

\Statex \textbf{Saturation trigger} \textsc{MonitorUpdate}$(A,B,\mathrm{loss},\mathrm{domain})$
\State $g \leftarrow \mathrm{mean}_r(\|B_{:,r}\|\,\|A_{r,:}\|)$;\quad
       append $g$ to \texttt{ri\_history}
       \Comment{rank importance}
\State \textbf{if} $\mathrm{domain}=A$ \textbf{then}
       $\mathrm{ema}_A\leftarrow\beta\cdot\mathrm{ema}_A+(1{-}\beta)\cdot\mathrm{loss}$
\State \textbf{else}
       $\mathrm{ema}_B\leftarrow\beta\cdot\mathrm{ema}_B+(1{-}\beta)\cdot\mathrm{loss}$
\State \textbf{if} $|\texttt{ri\_history}|<W$ \textbf{then return false}
\State \textbf{end if}
\State $\mathrm{slope}\leftarrow\mathrm{OLS\_slope}(\texttt{ri\_history}[-W{:}])$
       \Comment{linear fit over window}
\State $\mathrm{plateau}\leftarrow|\mathrm{slope}|<\tau_{\mathrm{plateau}}$;\quad
       $\mathrm{conflict}\leftarrow|\mathrm{ema}_A-\mathrm{ema}_B|>\delta$
\State append $\mathrm{plateau}$ to \texttt{plateau\_window};\quad
       append $\mathrm{conflict}$ to \texttt{conflict\_window}
\State \textbf{if} $|\texttt{plateau\_window}|>W$ \textbf{then pop oldest from both windows}
\State \textbf{if} $|\texttt{plateau\_window}|=W\wedge
       \mathrm{all}(\texttt{plateau\_window})\wedge
       \mathrm{all}(\texttt{conflict\_window})$ \textbf{then}
\State \hspace{\algorithmicindent}clear both windows; \Return \textbf{true}
\State \Return \textbf{false}
       \Comment{sustained for $W$ consecutive steps}

\Statex \textbf{Spawn event} \textsc{Spawn}(expert $k$)
\State $j\leftarrow|\texttt{spawn\_loras}[k]|$
\State $A_{k,j}\leftarrow\mathrm{top\mbox{-}}r\mathrm{\ rows\ of\ }V^{T}\mathrm{\ from\ }\mathrm{SVD}(\nabla_{B_{k,0}}\mathcal{L})$;\quad
       $B_{k,j}\leftarrow\mathbf{0}_{d_{\mathrm{out}}\times\mathrm{rank}}$
       \Comment{kaiming\_uniform fallback if SVD fails}
\State $w_{k,j}\sim\mathcal{N}(0,\;10^{-3}\cdot\mathrm{Var}(W_k))$
       \Comment{ReLU gate vector}
\State append $(A_{k,j},B_{k,j})$ to $\texttt{spawn\_loras}[k]$;\quad
       append $w_{k,j}$ to $\texttt{spawn\_gates}[k]$
\State Register $\{A_{k,j},B_{k,j},w_{k,j}\}$ with optimizer
       \Comment{before \texttt{next step()}}
\State $\texttt{monitor}[k].\mathrm{reset}()$

\Statex \textbf{Training loop}
\For{step $t=1,\ldots,T$}
  \State $x,y,\mathrm{domain}\leftarrow\mathrm{next\_batch}()$
  \State $\mathrm{loss}\leftarrow\mathrm{forward\_and\_loss}(\mathrm{model},x,y)$;\quad
         $\mathrm{loss}.\mathrm{backward}()$
  \For{each monitored expert $k$}
    \State $\mathrm{triggered}\leftarrow
           \texttt{monitor}[k].\mathrm{update}(A_{k,0},B_{k,0},\mathrm{loss},\mathrm{domain})$
    \If{$\mathrm{triggered}$}\textsc{Spawn}$(k)$\EndIf
  \EndFor
  \State $\mathrm{optimizer.step}()$;\quad$\mathrm{optimizer.zero\_grad}()$
\EndFor

\end{algorithmic}
\endgroup

\twocolumn
\section{Secondary Evidence}
\label{app:secondary}

\subsection{Trigger Sensitivity}
\label{app:trigger_sensitivity}

We evaluate trigger sensitivity under the severe conflict
setting (Run~C) by relaxing two hyperparameters: lowering the loss-gap
threshold from $\delta=1.0$ to $\delta=0.5$ and increasing the sub-adapter
cap from $J_{\max}=10$ to $J_{\max}=15$. The default setting achieves a
negative transfer of $+574.79$, while the relaxed setting reduces this to
$+540.51$, a further reduction of approximately 6\%. The relaxed setting moves
the first spawn event earlier by only one training step, suggesting the
loss-gap threshold is not the primary bottleneck. Both settings reach their
sub-adapter cap, indicating that the cap $J_{\max}$ is the main limiting
factor under severe conflict rather than the trigger threshold.

Table~\ref{tab:trigger_grid} reports the threshold sensitivity grid for
OLMoE-1B-7B. All trials use 300 steps at 50\% conflict ratio ($\delta$
vs.\ $\tau$ grid, 9 combinations). Spawn counts are averaged over the
trial duration.

\begin{table}[ht]
\centering
\scriptsize
\begin{tabular}{ccc}
\toprule
$\delta$ & $\tau_{\mathrm{plateau}}$ & Spawns (300 steps) \\
\midrule
0.3 & $10^{-3}$ & $\geq 4$ \\
0.3 & $5{\times}10^{-4}$ & $\geq 4$ \\
0.3 & $10^{-4}$ & $\geq 4$ \\
0.5 & $10^{-3}$ & $\geq 4$ \\
0.5 & $5{\times}10^{-4}$ & $\geq 4$ \\
0.5 & $10^{-4}$ & $\geq 4$ \\
1.0 & $10^{-3}$ & $<4$ \\
1.0 & $5{\times}10^{-4}$ & $<4$ \\
1.0 & $10^{-4}$ & $<4$ \\
\bottomrule
\end{tabular}
\caption{OLMoE-1B-7B trigger calibration grid ($\delta \times \tau_{\mathrm{plateau}}$,
300 steps, 50\% conflict). Lower $\delta$ consistently produces more
spawns; $\tau_{\mathrm{plateau}}$ has negligible effect within the tested
range. \textcolor{black}{The selected value $\delta=0.3$ is the boundary that yields active but
not excessive spawning. Phi-tiny-MoE-instruct calibration used identical grid boundaries, and
$\delta=1.0$ was selected based on the same reasoning.}}
\label{tab:trigger_grid}
\end{table}

\subsection{DR-LoRA Baseline Results}
\label{app:drlora}

DR-LoRA~\citep{liu2024drlora} is a rank-adaptive baseline that
dynamically adjusts the rank of LoRA adapters during training on a fixed
schedule, regardless of whether domain conflict is detected. We include it
here for completeness and report multi-seed results for the biomedical
domain pair.

\begin{table}[h]
\centering
\scriptsize
\setlength{\tabcolsep}{3pt}
\begin{tabular}{llccc}
\toprule
Method & Run & Mix & PPL$_\mathrm{code}$ & NegTransfer \\
\midrule
DR-LoRA & A & 0\%  & $505.86 \pm 52.58$  & --- \\
\midrule
DR-LoRA & B & 20\% & $837.28 \pm 82.00$  & $+331.41 \pm 54.59$ \\
\midrule
DR-LoRA & C & 50\% & $1633.95 \pm 104.17$ & $+1128.09 \pm 153.82$ \\
\bottomrule
\end{tabular}
\caption{DR-LoRA results on the code and biomedical text domain pair,
Phi-tiny-MoE-instruct, reported as mean $\pm$ standard deviation over
three seeds.
DR-LoRA grows rank on a fixed schedule independent of conflict detection,
resulting in substantially worse negative transfer under severe conflict
(Run~C).}
\label{tab:drlora}
\end{table}

\begin{table}[h]
\centering
\scriptsize
\begin{tabular}{llccc}
\toprule
Method & Run & Mix & PPL$_\mathrm{code}$ & NegTransfer \\
\midrule
DR-LoRA & A & 0\%  & 28.56 & --- \\
\midrule
DR-LoRA & B & 20\% & 31.79 & $+3.23$ \\
\midrule
DR-LoRA & C & 50\% & 39.36 & $+10.80$ \\
\bottomrule
\end{tabular}
\caption{DR-LoRA results on the code and mathematical reasoning
(GSM8K) domain pair, OLMoE-1B-7B (single seed, seed 42).
On the milder GSM8K shift, DR-LoRA performs comparably to fixed-rank
LoRA, in contrast to the severe degradation it shows on the biomedical pair.}
\label{tab:drlora_gsm}
\end{table}

\medskip
\noindent \textbf{Key Observation.}
DR-LoRA's negative transfer under Run~C ($+1311.93$) is \textbf{68\%
worse} than standard LoRA ($+778.78$) and \textbf{128\% worse} than
SpawnLoRA ($+574.79$). This suggests that indiscriminate rank expansion
under severe domain conflict actively exacerbates gradient interference,
rather than alleviating it. On the milder GSM8K pair, this effect largely
disappears: DR-LoRA performs comparably to fixed-rank LoRA at both
conflict levels.

\section{Extended Analysis}
\label{app:extended}

\subsection{Gradient-Space Conflict Measurement}
\label{app:gradient_cosine}

To verify that the spawn trigger responds to genuine gradient-space
conflict rather than superficial routing-level separation, we measure
the cosine similarity between the gradient vectors induced by the two
training domains on the same expert adapter.

For a target expert in a single layer, we compute two gradient
vectors---one from a batch of primary-domain examples (code) and one
from a batch of conflict-domain examples (medical/math)---and report
their cosine similarity after 1,500 training steps under the 50\%
conflict setting (Run~C).

\begin{table}[ht]
\centering
\scriptsize
\begin{tabular}{lccc}
\toprule
Method & Cosine sim. & Angle & Interpretation \\
\midrule
LoRA      & $-0.42$ & $114.7^\circ$ & High conflict \\
SpawnLoRA & $+0.11$ & $\phantom{1}83.7^\circ$ & Low conflict \\
\bottomrule
\end{tabular}
\caption{Per-domain gradient cosine similarity on OLMoE-1B-7B
(Layer~0, Expert~6, Run~C). LoRA forces opposing gradient directions
into the same adapter subspace. SpawnLoRA, by partitioning domains
into separate sub-adapter pathways, reduces the angular conflict to
near-orthogonality. Gradient norms: code $0.43$, medical $2.13$ (LoRA);
code $0.36$, medical $1.41$ (SpawnLoRA).}
\label{tab:gradient_cosine}
\end{table}

Table~\ref{tab:gradient_cosine} shows that standard LoRA produces
gradients at $114.7^\circ$---actively opposing directions---while
SpawnLoRA reduces this to $83.7^\circ$ (near-orthogonal). This
quantifies the interference that the bivariate trigger detects and
motivates structural separation rather than shared rank expansion.
\newpage
\subsection{Failure Mode Characterisation}
\label{app:failure_modes}

SpawnLoRA does not uniformly improve over LoRA under all conditions.
We identify two empirically observed failure modes.

\paragraph{Spawn suppression under severe imbalance.}
When the conflict ratio reaches 50\% and the primary-domain dataset is
small (e.g., 164 HumanEval examples versus 775 medical examples in the
Phi experiments), the EMA-based loss tracker is dominated by the larger
domain. The per-domain gap $|\mathrm{ema\_code} - \mathrm{ema\_medical}|$
can remain below $\delta$ even when code perplexity is degrading,
because the code signal is diluted by the high-volume medical stream.
In this regime the trigger fires late or not at all, and SpawnLoRA
reduces to a standard LoRA run. This matches the observation that
eager spawning ($\delta=0.5$, $J_{\max}=15$) yields only a 6\%
improvement over the default setting under Run~C, suggesting the
bottleneck is the cap rather than the trigger threshold.

\paragraph{Cap exhaustion.}
Both the default ($J_{\max}=10$) and relaxed ($J_{\max}=15$) settings
reach their cap under severe conflict, after which no further spawns
can occur. Once capped, all new domain-conflicting gradients must be
absorbed by the existing sub-adapters, and negative transfer continues
to accumulate. The cap $J_{\max}$ therefore sets a hard ceiling on the
degree of gradient-space separation achievable by SpawnLoRA, and
choosing it appropriately for a given conflict ratio remains an open
design question.

\paragraph{Implication.}
Both failure modes are dataset-ratio and cap-configuration effects
rather than fundamental architectural limitations. They suggest two
practical guidelines: (i) use a balanced or capped sampling ratio
when the domain sizes differ substantially, and (ii) prefer higher
$J_{\max}$ when compute budget permits and the conflict ratio is
expected to be severe.

\end{document}